\pdfoutput=1

\documentclass[11pt]{article}

\usepackage[]{acl}

\usepackage{times}
\usepackage{latexsym}
\usepackage{tcolorbox}
\usepackage{caption}
\usepackage{float}
\usepackage{subcaption}
\usepackage{booktabs}
\usepackage{multirow}
\usepackage{chngcntr}
\usepackage{hhline} 
\usepackage{tabularx}
\usepackage{highlight}
\usepackage{colortbl}
\usepackage{amsmath}

\usepackage[T1]{fontenc}

\usepackage[utf8]{inputenc}

\usepackage{microtype}

\title{
\CodeForStruct: Code Generation for Few-Shot Event Structure Prediction
}

\newcommand{\eg}{{\sl e.g.}}
\newcommand{\ie}{{\sl i.e.}}

\newcommand{\sref}[1]{\S\ref{#1}}
\newcommand{\fref}[1]{Figure~\ref{#1}}
\newcommand{\tref}[1]{Table~\ref{#1}}

\newcommand{\codex}{{\textsc{Codex}}}
\newcommand{\CodeForStruct}{{\textsc{Code4Struct}}}

\author{
    Xingyao Wang \and Sha Li \and Heng Ji \\
    University of Illinois Urbana-Champaign, IL, USA \\
    \texttt{\{xingyao6, shal2, hengji\}@illinois.edu}
}

\begin{document}
\maketitle
\begin{abstract}

Large Language Model (LLM) trained on a mixture of text and code has demonstrated impressive capability in translating natural language (NL) into structured code.
We observe that semantic structures can be conveniently translated into code and propose \CodeForStruct~to leverage such text-to-structure translation capability to tackle structured prediction tasks.
As a case study, we formulate Event Argument Extraction (EAE) as converting text into event-argument structures that can be represented as a class object using code. 
This alignment between structures and code enables us to take advantage of Programming Language (PL) features such as inheritance\footnote{\label{footnote:inheritance}Inheritance is a way to create a hierarchy of classes in PL. A child class can base upon another class, retaining similar implementation.} and type annotation\footnote{\label{footnote:type-annotation}Developers use type annotations to indicate the data types of variables and input/outputs of functions.} to introduce external knowledge or add constraints.
We show that, with sufficient in-context examples, formulating EAE as a code generation problem is advantageous over using variants of text-based prompts.
Despite only using 20 training event instances for each event type, \CodeForStruct\ is comparable to supervised models trained on 4,202 instances and outperforms current state-of-the-art (SOTA) trained on 20-shot data by 29.5\% absolute F1. 
%
By leveraging the inheritance feature of PL,
\CodeForStruct~can use 10-shot training data from a sibling event type to predict arguments for zero-resource event types and outperforms the zero-shot baseline by 12\% absolute F1.
\footnote{All code and resources are publicly available at \url{https://github.com/xingyaoww/code4struct}.}

\end{abstract}
\begin{figure}[h!]
\centering
\includegraphics[width=0.5\textwidth]{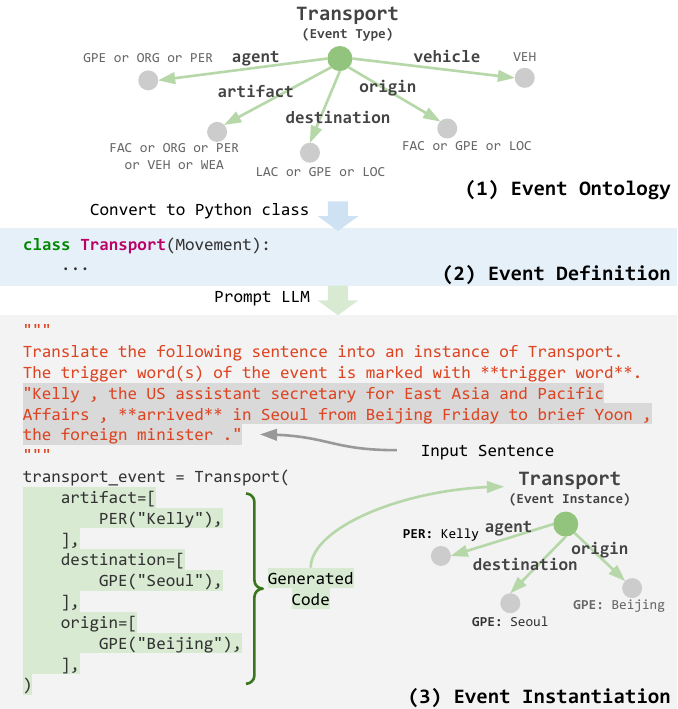}
\caption{Event Argument Extraction using code generation. 
We convert the existing event type ontology to \textsc{Python} class definitions. Conditioned on these definitions, we put the input sentence for event argument extraction into a docstring as the prompt for code generation. The generated code (colored in green) can be mapped to an instance graph of \texttt{Transport} event.
}
\label{fig:EAE-as-codegen-problem}
\end{figure}

\section{Introduction}
\begin{table*}[t!]
\centering
\includegraphics[width=0.95\textwidth]{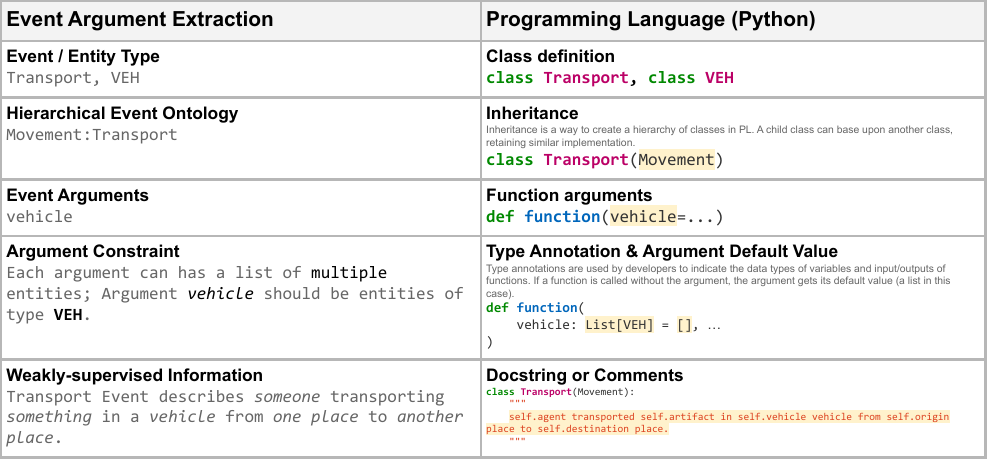}
\caption{Mapping between Event Argument Extraction requirements and features of Python programming language.
}
\label{tab:EAE-PL-Comparison}
\end{table*}

Large Language Model (LLM) trained on massive corpora of code mixed with natural language (NL) comments and docstrings\footnote{Text used to document a specific segment of code.} (\eg, \citealt{codex}, \citealt{CodeGenNijkamp2022ACP}) has demonstrated the ability to translate natural language instructions into structured code.
%
We ask if this conversion between language and code can serve as a bridge to build a connection between language and semantic structure, which is the goal of many structured prediction tasks (\eg, semantic parsing, information extraction) in Natural Language Processing (NLP). 
In particular, the target structure (\eg, event-argument graph in \fref{fig:EAE-as-codegen-problem}) can be mapped to code more straightforwardly compared to natural language, which often requires careful prompt engineering (\citealt{hsu-etal-2022-degree}, \citealt{li-etal-2021-document}, \tref{tab:example-EAE-language-template}). 
In addition, code written in programming languages has an inherent advantage in representing complex and interdependent structures~\cite{nl_for_pl_5387906, Sebrechts1985ProgrammingIN} with features such as inheritance and type annotation.

As a case study, we showcase our proposed \CodeForStruct\ on the Event Argument Extraction (EAE) task, which aims to extract event structures from unstructured text. EAE is the ideal testbed for our method due to the close alignment between EAE and PL as shown in \tref{tab:EAE-PL-Comparison}.
In \CodeForStruct\ (\fref{fig:EAE-as-codegen-problem}), we first translate the entity and event type ontology into Python class definitions. Conditioned on the relevant class definitions and the input sentence, we prompt an LLM to generate an instantiation of the event class, from which we can extract the predicted arguments. 

By leveraging the alignment between PL and NLP problems, \CodeForStruct\ enjoys various advantages as shown in \tref{tab:EAE-PL-Comparison}.
Using PL features like type annotation and argument default value, we can naturally enforce argument constraints for output structures. This allows \CodeForStruct\ to handle multiple or zero argument fillers for the same argument role by annotating the expected type (\ie, expect a list of entities) and setting the default value for each argument (\ie, an empty list without any entity by default).
Furthermore, we can naturally utilize the event hierarchy by leveraging inheritance. Inheritance allows a child event class (\eg, \texttt{Transport}) to reuse most components of its parent class (\eg, \texttt{Movement}) while preserving its unique properties. 
We demonstrate that hierarchical event types allow zero-resource event types to use annotated training examples from their high-resource sibling types (\sref{sec:hierarchy-improve-low-resource-EAE}).

We outline our contributions as follows:
\begin{itemize}
    \item We propose \CodeForStruct\ to tackle structured prediction problems in NLP using code generation. As a case study, we use \CodeForStruct~for Event Argument Extraction (EAE).
    
    \item We perform extensive experiments contrasting the performance of code-based prompt and two variants of text prompt on different LLMs and show that code prompt is generally advantageous over text prompt when sufficient in-context examples are provided (\sref{sec:compare_textprompt}).
    
    \item We demonstrate that 20-shot \CodeForStruct\ rivals fully-supervised methods trained on 4,202 instances. \CodeForStruct~outperforms a SOTA approach by 29.5\% absolute F1 gain when 20-shot data are given to both. 0-shot \CodeForStruct~can even outperform the SOTA on both 20 and 50 shots (\sref{sec:performance-comparison}).
    
    \item We show that integrating the event ontology hierarchy by class inheritance can improve prediction. Compared to the zero-shot baseline, we see 12\% F1 gains for zero-resource event types when using 10-shot examples from their sibling event types (\sref{sec:hierarchy-improve-low-resource-EAE}).
\end{itemize}

\begin{figure*}[th!]
\centering
\includegraphics[width=\textwidth]{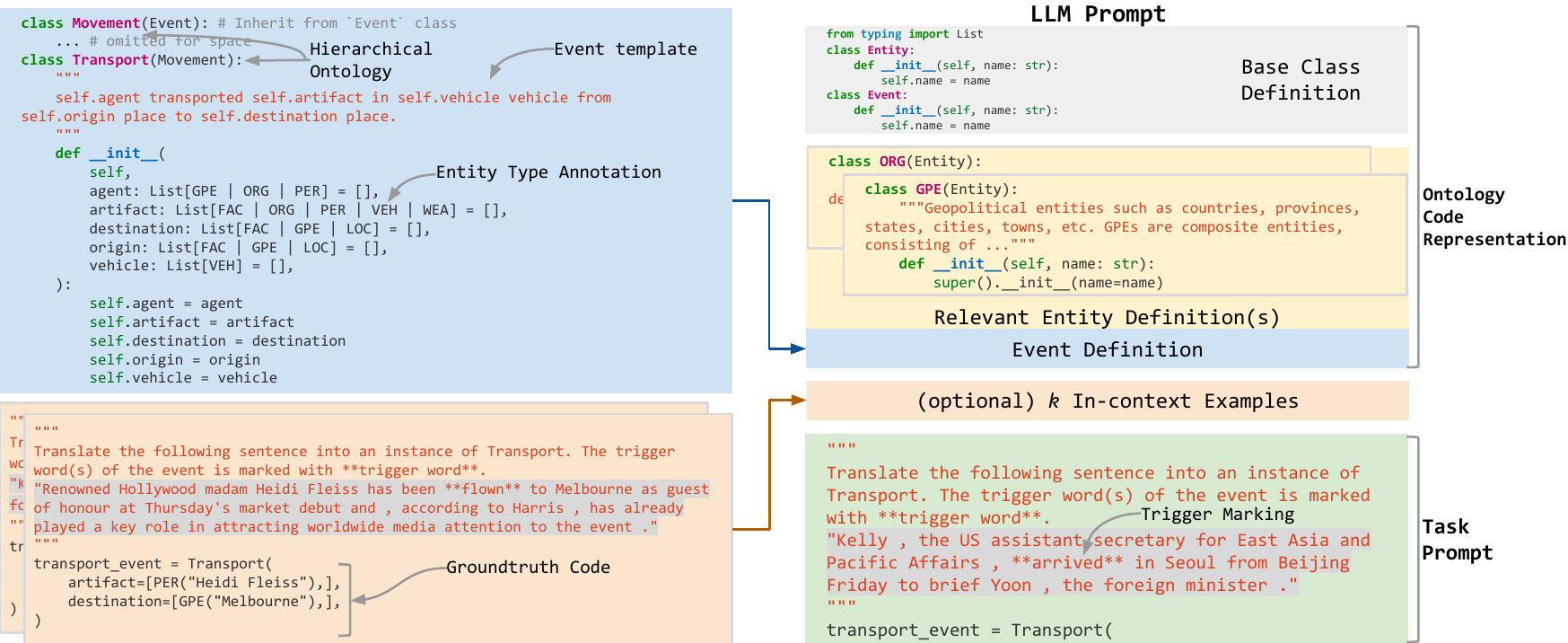}
\caption{Prompt components. \textbf{(1)} Ontology code representation contains definitions of entity and event classes, colored in yellow and blue (\sref{sec:ontology-construction}). \textbf{(2)} $k$-shot examples for in-context learning, colored in orange (\sref{sec:in-context-learning}). \textbf{(3)} The task prompt, appended at the end with partial class instantiation for LLM completion, colored in green (\sref{sec:task-prompt}).}
\label{fig:prompt-breakdown}
\end{figure*}

\section{Code Generation Prompt Construction}
\label{sec:prompt-construction}
In Event Argument Extraction (EAE) task, a model is provided with an event ontology and the target text to extract from. 
Similarly, we prompt an LLM with the ontology that consists of definitions of event types and argument roles, and input sentences to generate code that instantiates the given event type. 
We breakdown the input prompt into three components: (1) ontology code representation which consists of Python class definitions for entity types and an event type (\sref{sec:ontology-construction}); (2) optional \textit{k}-shot in-context learning examples for the event type defined in (1) (\sref{sec:in-context-learning}); (3) task prompt for completion (\sref{sec:task-prompt}). We show a breakdown of the full prompt in \fref{fig:prompt-breakdown}. 

\subsection{Ontology Code Representation}
\label{sec:ontology-construction}
To represent the event ontology as code, we concatenate the base class definition, entity class definitions, and event class definitions. 

\paragraph{Base Class Definition} We define base type \texttt{Entity} and \texttt{Event} to be inherited by other classes.

\paragraph{Entity Class Definition} We use entity type definitions from the Automatic Content Extraction (ACE) program\footnote{\url{https://www.ldc.upenn.edu/collaborations/past-projects/ace}}.  
We construct Python classes that inherit from \texttt{Entity} and use the entity type as the class name (\eg, \texttt{class GPE(Entity)}). We add a natural language description as a docstring of the defined class for each entity type.

\subsubsection{Event Class Definition} 

We define the event class using the name of the event type (\eg, \texttt{class Transport}). 
As ACE defines its event types in a hierarchical ontology, mimicking class definitions in Object-Oriented PL, we inherit the event class definition from its parent (\eg, \texttt{class Transport(Movement)}) or root event type if the event class does not has a parent (\eg, \texttt{class Movement(Event)}).  An example of hierarchical event definition can be found in \fref{fig:example-hierarchical-event-definition}. 

We define the argument roles (\eg, destination of \texttt{Transport}) as input arguments of the constructor \texttt{\_\_init\_\_}\footnote{A constructor is a special function that initializes an instance of a class.}. We specify the type of each argument role using Python type annotation, a commonly used PL feature: For example, \texttt{agent: List[GPE | ORG | PER]} means that the \texttt{agent} argument accepts a list of entities which could be either of type GPE (Geo-Political Entity), ORG (Organization), or PER (Person). 
We assign each input argument (\eg, \texttt{agent}) to a class member variable of the same name.

We include event description templates into the docstring of the class definition.
The event description templates are modified from \citet{li-etal-2021-document} by replacing each role with their corresponding member variable (\eg, \texttt{self.agent}).

\begin{table*}[th]
\centering
\resizebox{\textwidth}{!}{
\begin{tabular}{p{4.5cm}p{17cm}}
\toprule
Prior Work & Language Template \\
\midrule
DEGREE \cite{hsu-etal-2022-degree} & \textbf{somebody} was moved to \textbf{somewhere} from \textbf{some place} by \textbf{some way}. \textbf{somebody} or \textbf{some organization} was responsible for the movement.
\textbf{something} was sent to \textbf{somewhere} from \textbf{some place}. \textbf{somebody} or \textbf{some organization} was responsible for the \textbf{transport}. \\ \hline

BART-Gen \cite{li-etal-2021-document} & \textbf{<arg1>} \textbf{transported} \textbf{<arg2>} in \textbf{<arg3>} vehicle from \textbf{<arg4>} place to \textbf{<arg5>} place \\ \hline

Text2Event \cite{lu-etal-2021-text2event} & 
(~(\textbf{Transport} returned (Agent \textbf{<arg>}) (Artifact \textbf{<arg>}) (Destination \textbf{<arg>}) (Origin \textbf{<arg>}) (Vehicle \textbf{<arg>})~) \\
\bottomrule
\end{tabular}
}
\caption{
Example of language templates for Event Argument Extraction used by \citet{hsu-etal-2022-degree, li-etal-2021-document, lu-etal-2021-text2event}.
}
\label{tab:example-EAE-language-template}
\end{table*}

\subsection{Task Prompt}
\label{sec:task-prompt}

The task prompt consists of a docstring describing the task and incomplete event instantiation code for completion. An example of a task prompt can be found in \fref{fig:prompt-breakdown}.
The text-based docstring contains a task instruction and an input sentence. We mark the ground truth trigger words for the input text by surrounding them with \texttt{**}. We choose to use \texttt{**} as it is used to set text to bold in \texttt{Markdown} (a markup language for creating formatted text), which is commonly found in code bases and web data on which our LLM is trained. 
The incomplete code prompt assigns a partial instantiation of an event class to a variable to trigger the model for completion, for example, \texttt{transport\_event = Transport(}. 

We observed that LLM tends to generate additional sentences paired with extracted arguments if no stopping constraint is applied. To focus on the given EAE task, we stop the code generation whenever any of the following patterns is generated by the model: \texttt{"""}, \texttt{class}, \texttt{print}, or \texttt{\#}.

\subsection{In-context Learning}
\label{sec:in-context-learning}

Optionally, we can include in-context learning examples, which are task prompts (\sref{sec:task-prompt}) paired with completed event instantiations using ground-truth arguments (see \fref{fig:prompt-breakdown} for a specific example). 
For $k$-shot learning, we concatenate $k$ such examples together.
Given a task prompt, we deterministically gather $k$ learning examples by collecting training instances with the same event type, following the order of occurrences in the training set.

\section{Why Represent Event Structure in PL?}
\label{sec:problem-alignment}

A wide range of NLP tasks have benefited from LLM \cite{gpt3, hoffmann2022training, chowdhery2022palm} trained on web-scale language corpora.
To effectively use LLM trained on language for EAE, one of the biggest challenges is to specify the desired output, namely event structures in our case, using natural language.

There is a tradeoff between the effort put into defining the output or designing the prompt (\eg, Text2Event in \tref{tab:example-EAE-language-template}) and the benefit from pretraining in natural language (\eg, DEGREE and BART-Gen in \tref{tab:example-EAE-language-template}). 
Text2Event~\cite{lu-etal-2021-text2event} resides at one end of the spectrum with a concise but unnatural output format. As a result, this formulation under-utilizes the pretraining power of the model and does not work in low-resource settings as shown in \tref{tab:performance-comparison}. 
Towards the other end, \citet{hsu-etal-2022-degree,li-etal-2021-document} design manual templates for the model to fill in. We also design two variants of language prompt as shown in~\fref{fig:IE-puretext-gen} and \ref{fig:IE-naturaltext-gen} miciking our code prompt and BART-Gen style prompt for comparison. Note that these natural language prompts are much more verbose and, as shown in \sref{sec:compare_textprompt}, usually result in sub-optimal performance with sufficient in-context examples. 

Essentially, this tradeoff is a result of the mismatch between the pretraining corpora and task output formats. Instead of using LLM trained on only unstructured text, we turn to LLM trained with a mixture of text and code, where the text is often aligned in semantics with the accompanying code. Such Code-LLMs have the ability to convert text into corresponding code as demonstrated by \cite{codex,CodeGenNijkamp2022ACP}. Then we can map the desired output event structure into code in a straightforward manner and leverage the full pretraining power of these models.
PLs like Python offer features (\eg, class, docstrings, type annotations, inheritance) that have a significant presence in the pre-training corpus of Code-LLM due to frequent usage. \CodeForStruct\ leverages these features to succinctly describe event structures, which makes it better aligned with Code-LLM.
By leveraging LLM's learned knowledge from diverse pre-training domains, \CodeForStruct\ can work well in open-domain, achieving non-trivial zero-shot performance given unseen event types (\sref{sec:performance-comparison}).
\CodeForStruct\ is also data-efficient as exemplified by reaching comparable performance to fully-supervised methods with much fewer annotated examples (20 per event type) (\sref{sec:performance-comparison}).

\section{Experiments}

\label{sec:experiments}
\subsection{Experiment Setup}
\label{sec:experiment-setup}
\paragraph{LLM} We use \codex\ \texttt{code-davinci-002} \cite{codex}, a GPT-3 \cite{gpt3} model finetuned on code, which supports up to 8k input tokens. 
We compare its performance with InstructGPT \cite{Ouyang2022TrainingLM} \texttt{text-davinci-002} and its improved version \texttt{text-davinci-003}, both support up to 4k input tokens. 
We access these LLMs through OpenAI API\footnote{\url{https://openai.com/api/}}.

\paragraph{Hyperparameters} We prompt LLM to generate code that instantiates  an event using sampling temperature $t=0$ (\ie, greedy decoding).
We set the max number of new tokens for each generation to 128, which fits all code outputs for the test set.

\paragraph{Evaluation Tasks} We use ground truth event type and gold-standard trigger words to perform Event Argument Extraction. 

\paragraph{Dataset} We evaluate our performance of EAE on the English subset of Automatic Content Extraction 2005 dataset (ACE05-E)\footnote{\url{https://www.ldc.upenn.edu/collaborations/past-projects/ace}} \cite{doddington-etal-2004-automatic}. We follow \citet{wadden-etal-2019-entity, lin-etal-2020-joint} for dataset processing. 
ACE05-E has hierarchical event types with 8 parent types and 33 child types. Among all child types, roughly half of the event types (14 out of 33) in ACE05-E have less than 50 event instances in the training set. We show statistics for each event type in \tref{tab:event-instance-stats}.

\paragraph{Evaluation metrics} We use \textbf{Argument F1-score} following prior work \cite{ji2008refining,li-etal-2021-document, hsu-etal-2022-degree}:
We consider an argument to be correctly identified when the head word span of predicted text\footnote{We find the span of predicted text in the given sentence, then use \texttt{spacy} library to find its head word.} matches that of the human-annotated text (denoted as \textbf{Arg-I}); 
We consider an argument to be correctly classified if the role (\eg, \texttt{agent}) of a \emph{correctly identified} argument matches that of the human annotation (denoted as \textbf{Arg-C}).

\begin{table*}[t]
\centering
\caption{Performance of the code prompt on the Arg-I and Arg-C metrics and its F1 score difference $\Delta_{C-T}^{(i)}$ with two text prompt variants described in \sref{sec:compare_textprompt} (\ie, $\text{F1}_{\text{code}} - \text{F1}_{\text{text}}^{(i)}$).
On Arg-C, there is a trend that the code prompt performs better (\ie, $\Delta_{C-T}^{(i)} > 0$) when more in-context examples are provided, except on \texttt{text-davinci-003}.
}
\label{tab:text-vs-code-prompt}
\resizebox{\textwidth}{!}{

\begin{tabular}{r||ccc|ccc||ccc|ccc||ccc|ccc}
\toprule

Model & \multicolumn{6}{c}{code-davinci-002} & \multicolumn{6}{c}{text-davinci-002} & \multicolumn{6}{c}{text-davinci-003} \\
$k$-shot &            Arg-I & $\Delta_{C-T}^{(1)}$ & $\Delta_{C-T}^{(2)}$ & Arg-C & $\Delta_{C-T}^{(1)}$ & $\Delta_{C-T}^{(2)}$ &            Arg-I & $\Delta_{C-T}^{(1)}$ & $\Delta_{C-T}^{(2)}$ & Arg-C & $\Delta_{C-T}^{(1)}$ & $\Delta_{C-T}^{(2)}$ &            Arg-I & $\Delta_{C-T}^{(1)}$ & $\Delta_{C-T}^{(2)}$ & Arg-C & $\Delta_{C-T}^{(1)}$ & $\Delta_{C-T}^{(2)}$ \\
\midrule
0      &             50.6 &      \gcdiffa{0.7} &     \gcdiffb{50.6} &  36.0 &     \gcdiffa{-2.2} &     \gcdiffb{36.0} &             48.9 &     \gcdiffa{-2.6} &     \gcdiffb{20.2} &  35.0 &     \gcdiffa{-2.4} &     \gcdiffb{13.1} &             49.9 &     \gcdiffa{-2.1} &     \gcdiffb{15.3} &  37.8 &     \gcdiffa{-1.4} &     \gcdiffb{12.6} \\
1      &             57.3 &      \gcdiffa{0.1} &      \gcdiffb{4.7} &  47.8 &     \gcdiffa{-1.0} &      \gcdiffb{4.7} &             55.8 &      \gcdiffa{1.8} &      \gcdiffb{5.3} &  45.2 &      \gcdiffa{3.0} &      \gcdiffb{4.9} &             56.0 &     \gcdiffa{-1.5} &      \gcdiffb{1.1} &  44.7 &     \gcdiffa{-3.2} &      \gcdiffb{1.1} \\
5      &             58.0 &      \gcdiffa{1.1} &      \gcdiffb{1.9} &  52.5 &      \gcdiffa{2.9} &      \gcdiffb{1.1} &             56.0 &     \gcdiffa{-2.0} &      \gcdiffb{1.0} &  48.8 &      \gcdiffa{3.0} &      \gcdiffb{1.4} &             59.2 &     \gcdiffa{-0.9} &     \gcdiffb{-0.7} &  51.7 &      \gcdiffa{1.4} &     \gcdiffb{-2.1} \\
10     &             57.2 &     \gcdiffa{-1.4} &     \gcdiffb{-0.2} &  52.8 &      \gcdiffa{0.8} &      \gcdiffb{0.1} &             60.6 &      \gcdiffa{2.7} &      \gcdiffb{2.9} &  53.9 &      \gcdiffa{6.4} &      \gcdiffb{5.0} &             62.8 &      \gcdiffa{3.1} &      \gcdiffb{0.6} &  56.3 &      \gcdiffa{5.0} &     \gcdiffb{-1.2} \\
20     &             62.1 &      \gcdiffa{1.7} &      \gcdiffb{0.2} &  58.5 &      \gcdiffa{3.6} &      \gcdiffb{2.4} &             59.9 &      \gcdiffa{0.9} &      \gcdiffb{3.7} &  56.5 &      \gcdiffa{8.0} &      \gcdiffb{5.8} &             65.0 &      \gcdiffa{3.5} &      \gcdiffb{0.7} &  60.4 &      \gcdiffa{7.8} &     \gcdiffb{-0.4} \\

\bottomrule

\end{tabular}
}
\end{table*}

\subsection{Comparison with Text Prompt}
\label{sec:compare_textprompt}

To compare our code-based prompt with text-based prompts, we design two variants of text prompt: $T^{(1)}$ mimicking our code prompt (\ie, code imitation, \fref{fig:IE-puretext-gen}) and $T^{(2)}$ following BART-Gen style prompt \cite{li-etal-2021-document} (\fref{fig:IE-naturaltext-gen}) which resembles natural language more compared to $T^{(1)}$.
Both text prompts have similar components as our code-based prompt in \fref{fig:prompt-breakdown}. 
Text prompts rely on natural language to define the requirement and format of the desired output, while the code prompt utilizes PL syntax.
%
We compare the F1 score difference between the code prompt (\sref{fig:prompt-breakdown}) and two variants of text prompts (\ie, $\Delta_{C-T}^{(i)} = \text{F1}_{\text{code}} - \text{F1}_{\text{text}}^{(i)}, i \in \{1, 2\}$) on different LLMs in \tref{tab:text-vs-code-prompt}.
We include exact performance numbers of text prompts in \tref{tab:text-and-code-prompt-perf-numbers}.
We summarize our findings as follows:
\begin{itemize}
    \item Code prompt outperforms both text prompts on Arg-C F1 (\ie, $\Delta_{C-T}^{(i)} > 0$) for two text prompt variants and all LLMs except \texttt{text-davinci-003} when sufficient in-context examples are given (\ie, $k \geq 5$).

    \item For \texttt{*-davinci-002} LLMs, there are more significant performance gains from using a code prompt (\ie, increasing $\Delta_{C-T}^{(i)}$ for all $i$) when the number of in-context examples $k$ increases (for $k \geq 5$).

    \item There is no clear trend on Arg-I F1 to differentiate code and text prompts, except for \texttt{text-davinci-003}, which exhibits similar behavior that code prompt performs better with larger $k$.

    \item Text prompt $T^{(2)}$ (BART-Gen style), which resembles natural language more, performs poorly under low-shot ($k \leq 1$), primarily due to the LLM being unable to produce the desired structure output described using language in $T^{(2)}$, causing the low-shot code-text performance gap $\Delta_{C-T}^{(2)}$ to be larger compared to $T^{(1)}$. These low-shot performance differences between $T^{(1)}$ and $T^{(2)}$ further signify the need to prompt engineering for language-based prompts to work well in a low-shot setting.

\end{itemize}

\subsection{Comparison with different LLM}

\begin{figure}[h]
\centering
\resizebox{\columnwidth}{!}{
\includegraphics[]{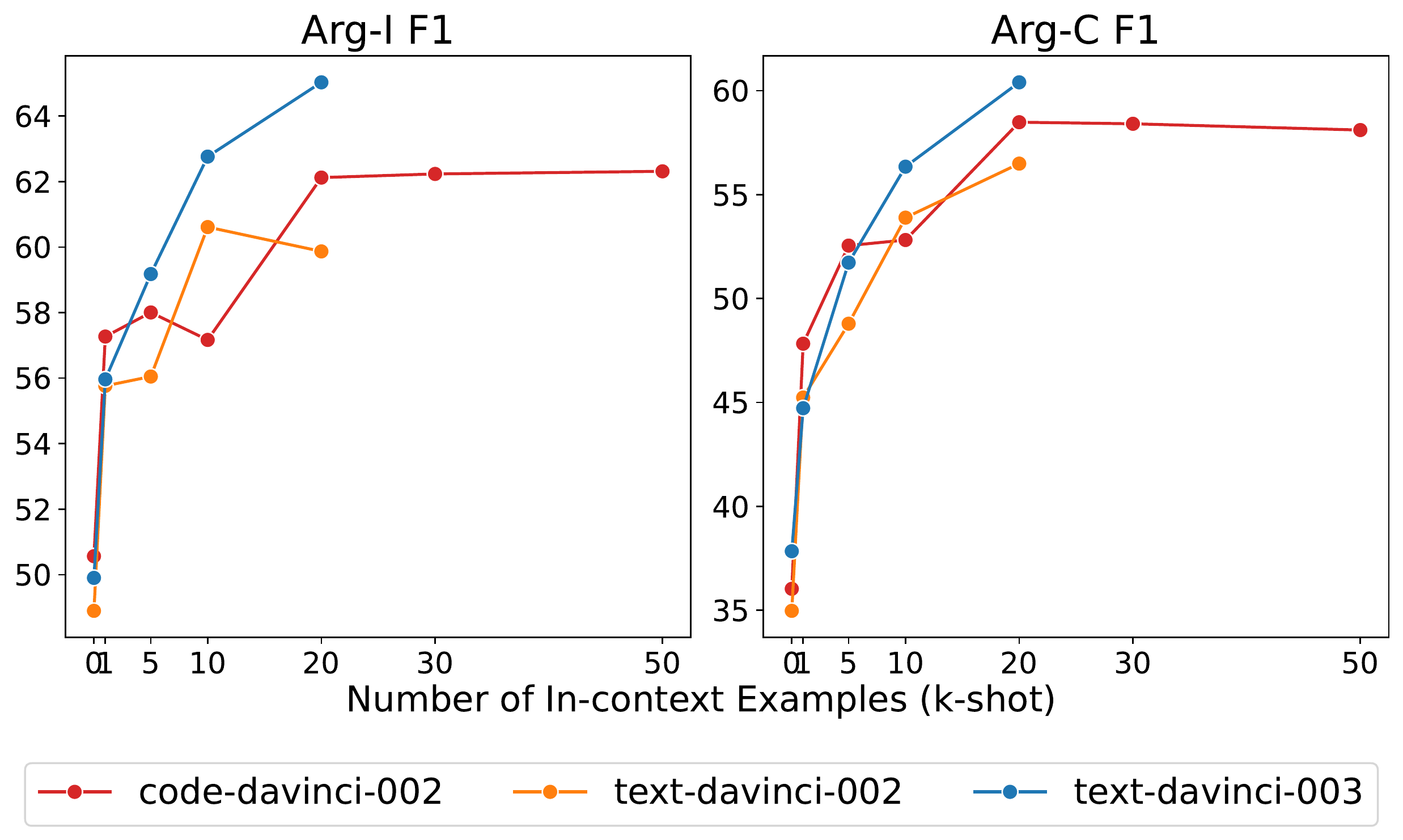}
}
\caption{
\CodeForStruct\ performance (in F1\%) with different $k$. We observe improvements with diminishing returns when we increase the number of in-context examples.
Exact performance numbers can be found in \tref{tab:text-and-code-prompt-perf-numbers} (code prompt).
We stop at $k=20$ for \texttt{text-davinci} and $k=50$ for \texttt{code-davinci-002} as including more examples would exceed the input length limitation imposed by corresponding LLM.
}
\label{fig:model-performance}
\end{figure}

We measure the performance of the same \CodeForStruct\ code prompt across different foundational LLMs in \sref{sec:experiment-setup}. LLM performance comparison can be found in \fref{fig:model-performance}.
\texttt{text-davinci-002} is an InstructGPT \cite{Ouyang2022TrainingLM} model finetuned with human demonstrations based on \texttt{code-davinci-002}, yet these two LLMs perform similarly in Arg-C F1. 
Although having a similar code prompt Arg-C performance, \texttt{text-davinci-002} generally has a larger $\Delta_{C-T}^{(i)}$ compared to \texttt{code-davinci-002} of the same $k$ in \tref{tab:text-vs-code-prompt} (\eg, +3.6 vs. +8.0, +2.4 vs. +5.8 on 20-shot for both text prompt variants), suggesting the degradation of text prompt performance after finetuning with human demonstrations. 

\texttt{text-davinci-003}, which uses reinforcement learning (RL) with reward models to align with human preference\footnote{\url{https://beta.openai.com/docs/model-index-for-researchers}} \cite{Ouyang2022TrainingLM}, outperforms other LLMs for $k > 5$.
In \tref{tab:text-vs-code-prompt}, \texttt{text-davinci-003} obtains superior Arg-C F1 performance (60.4\% vs. 56.5\% on 20-shot) compared to \texttt{text-davinci-002}. This suggests RL with reward models effectively improves EAE performance (\ie, Arg-C) on code prompt.

Interestingly, \texttt{text-davinci-003} has a very different $\Delta_{C-T}^{(i)}$ pattern for text prompt $T^{(2)}$ compared to $T^{(1)}$. Like \texttt{text-davinci-002}, in \tref{tab:text-vs-code-prompt}, Arg-C $\Delta_{C-T}^{(1)}$ for text prompt $T^{(1)}$ has an increasing trend with a similar magnitude (\eg, $+7.8$ vs. $+8.0$ on 20-shot). That is, in both LLMs, the code prompt is always better than text prompt $T^{(1)}$ with $k \geq 5$.
However, for text prompt $T^{(2)}$ which is more similar to natural sentences, the gap $\Delta_{C-T}^{(2)}$ exhibits a vastly different pattern compared to other models: code prompt performs on par or even slightly worse than $T^{(2)}$ for $k \geq 5$. We also notice that for zero-shot prediction, $T^{(2)}$ on \texttt{text-davinci-003} performs better compared to other LLMs. This indicates that aligning LLM with RL and reward models helps improve LLM's ability to follow zero-shot language instructions.

Even though code prompt still performs superior to both text prompt variants on \texttt{002} LLMs, results from \texttt{text-davinci-003} suggest a better-aligned language model can perform equally well on a natural text prompt $T^{(2)}$ when sufficient in-context examples are provided.

\subsection{Comparison with different $k$}

We examine the performance of code prompts with varying numbers of examples in \fref{fig:model-performance}.
We observe that F1 scores for all metrics generally increase with diminishing returns when providing more in-context learning examples.
The initial in-context example ($k=1$) brings the largest absolute performance gain (+11.8, +10.2, +6.9 Arg-C F1 for three LLMs). 
For $k\geq20$ on \texttt{code-davinci-002}, the Arg-I and Arg-C performance plateaus or even slightly degrade, as not all event types have enough in-context examples to benefit from increasing $k$ (\ie, only 19 out of 33 event types have more than 50 examples for in-context learning).
To further investigate why the performance plateaus, we analyze how the sentence variability (or diversity) of in-context examples influences Arg-C performance in \sref{sec:variability-analysis}; We find that Arg-C performance is positively correlated with the variability of in-context examples which plateaus as $k$ increases, hinting that in-context learning performance may eventually plateau with increasing $k$ due to little variability gains from the additional data.

\subsection{Comparison with Supervised Models}
\label{sec:performance-comparison}

\paragraph{Baselines}
Unlike prior methods trained on the entire training set, \CodeForStruct\ learns from up to 50 examples (\ie, 39 examples per event type on average, roughly 1\% among all training instances) to predict arguments for each test event type. 
To ensure a fair comparison, for each event type \textit{t} in the test set, we train a Text2Event model \cite{lu-etal-2021-text2event} and a DEGREE model (SOTA, \citet{hsu-etal-2022-degree}) on 20-shot and 50-shot in-context examples \CodeForStruct~used while providing gold-standard trigger words. 
We evaluate both models trained on event type \textit{t} on a partition of the test set that only contains instances of event type \textit{t}. We then aggregate F1 scores (micro F1) across all 31 event types on the test set and report them in \tref{tab:performance-comparison}.
Following \citet{hsu-etal-2022-degree}, we also compare with classification-based (DyGIE++ \citet{wadden-etal-2019-entity}, BERT\_QA \citet{du-cardie-2020-event}, OneIE \citet{lin-etal-2020-joint}) or generation-based (TANL \cite{paolini2021structured}, BART-Gen \citet{li-etal-2021-document}, DEGREE \citet{hsu-etal-2022-degree}) models trained on the full training set.

\paragraph{Results}
We report the performance of \CodeForStruct\ using LLMs (\sref{sec:experiment-setup}) in comparison with prior work in \tref{tab:performance-comparison}. We report the performance of supervised models using the full dataset from \citet{hsu-etal-2022-degree}. Note that 50-shot results for \texttt{text-davinci} are not available as the 50-shot input prompt will exceed LLM's input token length limitation, hence we use \texttt{code-davinci-002} for 50-shot comparison.

In the few-shot setting, 20-shot \CodeForStruct\ using \texttt{text-davinci-003} can surpass DEGREE \cite{hsu-etal-2022-degree}, the current state-of-the-art, by a large margin (+29.5\% Arg-C F1).
Our zero-shot \CodeForStruct\ using the best-performing \texttt{text-davinci-003} model can already achieve higher Arg-I and Arg-C performance than the 20-shot and 50-shot DEGREE. 
Despite only learning from 20 examples, 20-shot \CodeForStruct\ achieves comparable performance with other fully-supervised models trained on 100\% of the training data (4,202 instances).

\begin{table}[h!]
\centering
\resizebox{\columnwidth}{!}{%
\begin{tabular}{llcc}
\toprule
Model & Data & Arg-I F1 & Arg-C F1\\
\midrule
DyGIE++ & Full & 66.2 & 60.7 \\
BERT-QA & Full & 68.2 & 65.4 \\
OneIE   & Full & 73.2 & 69.3 \\
TANL    & Full & 65.9 & 61.0 \\
BART-Gen & Full & 69.9 & 66.7 \\
DEGREE  & Full & \textbf{76.0} & \textbf{73.5} \\ 
\hline
$\textbf{\CodeForStruct}_{\small\texttt{text-davinci-003}}$ & 0-shot & 49.9 & 37.8 \\
\hline
Text2Event & 20-shot* & 23.1 & 19.1 \\
DEGREE & 20-shot* & 33.0 & 30.9 \\ 
$\textbf{\CodeForStruct}_{\small\texttt{text-davinci-003}}$ & 20-shot* & \textbf{65.0} & \textbf{60.4} \\
\hline
Text2Event & 50-shot* & 30.6 & 26.0 \\
DEGREE & 50-shot* & 40.8 & 37.3 \\
$\textbf{\CodeForStruct}_{\small\texttt{code-davinci-002}}$ & 50-shot* & \textbf{62.3} & \textbf{58.1} \\ 
\bottomrule
\end{tabular}%
}
\caption{Performance (in F1\%) comparison between best-performing \CodeForStruct~LLM and existing supervised approaches.
Performance numbers for all LLMs can be found in \tref{tab:text-and-code-prompt-perf-numbers}.
*Some event types do not have 20 or 50 examples for in-context learning; on average, we have 39 examples per type for a 50-shot prompt and 18 examples per type for 20-shot.
}
\label{tab:performance-comparison}
\end{table}

\subsection{Event Type Hierarchy Improves Zero-resource EAE}
\label{sec:hierarchy-improve-low-resource-EAE}

In this section, we show that \CodeForStruct, when provided with hierarchical event definitions and few-shot training instances $D_{e_s}$ from a sibling event type $e_s \in \text{Siblings}(e)$ under the same parent event type, can improve performance for child event type $e$ as good as if training instances $D_{e}$ from the same event type $e$ were used.
This allows zero-resource event types without annotated data to exploit the event type hierarchy and benefit from their high-resource siblings. 
We include an example task prompt with sibling examples in \fref{fig:example-hierarchical-sibling-event-example} and report our results in \tref{tab:performance-using-hierarchy-examples}.

\paragraph{Setup} We split the child types for each parent type into training and testing types by selecting the high-resource child type with the largest amount of training instances to be the training type and have the rest be testing types. The train-test split for ACE types can be found in \tref{tab:event-type-hierarchy-train-test-split}. 
Under the same parent event type, we use data instances from the training type (\ie, a sibling of testing types) as in-context examples to predict arguments for each testing type. We include event class definition (\fref{fig:prompt-breakdown}) for parent event type (\eg, \texttt{Transaction}), child training (sibling) event type (\eg, \texttt{Transfer\_Money}), and child testing event type (\eg, \texttt{Transfer\_Ownership}). We show an example of event definition with sibling type in \fref{fig:example-hierarchical-sibling-event-definition}.
The few-shot performance when using data from a sibling type $D_{e_s}$ is denoted with \textit{(sibling type)} in \tref{tab:performance-using-hierarchy-examples}.
To demonstrate the effectiveness of using data from sibling event types, we compare it with using training instances from the testing event type itself $D_{e}$ (denoted as \textit{(same type)}) and from a random non-sibling event type (denoted as \textit{(non-sibling type)}).

\begin{table}[h]
\centering
\resizebox{\columnwidth}{!}{
\begin{tabular}{lrr}
\toprule
{} &  Arg-I &  Arg-C \\
\midrule
0-shot                            &   52.8 &   42.9 \\
\hline
1-shot (same type)                &   54.3 &   50.2 \\
1-shot (sibling type)             &   \textbf{57.2} &   \textbf{51.9} \\
1-shot (non-sibling type)  &   56.3 &   50.3 \\
\hline
10-shot (same type)               &   58.7 &   \textbf{55.2} \\
10-shot (sibling type)            &   \textbf{60.8} &   54.9 \\
10-shot (non-sibling type) &   58.5 &   51.0 \\
\bottomrule
\end{tabular}
}
\caption{
\texttt{code-davinci-002} performance (in F1\%) when using examples from the same, sibling or non-sibling event types for in-context learning.
To ensure a fair comparison, F1 scores are aggregated from 23 test event types in \tref{tab:event-type-hierarchy-train-test-split} that contains more than 10 training instances.
}
\label{tab:performance-using-hierarchy-examples}
\end{table}

\paragraph{Results} We observe that \CodeForStruct, when prompted with training examples from sibling type, performs on par with the prompt that uses training examples from the testing type itself on 1-shot and 10-shot. 
The substantial performance gain (+9\% Arg-C F1 on 1-shot, +12\% Arg-C F1 on 10-shot, compared with 0-shot) contributed by sibling-type training examples demonstrate the potential of applying \CodeForStruct~to zero-resource event types with no training data by exploiting their hierarchical relationship with other high-resource event types.
Surprisingly, similar to the observation made by \citet{Min2022RethinkingTR}, using in-context examples from a random non-sibling type also benefits \CodeForStruct\ performance, albeit not as helpful as sibling examples under 10-shot.

\section{Related Work}

\paragraph{Code-LLM for Structured Task} 
\citet{sun2019program, progprompt} focus on procedural tasks that aim to control situated agents in an embodied environment by representing the procedure plan in code. \citet{madaan2022language} uses Code-LLM to generate a structured commonsense reasoning graph represented in code, which is similar in spirit to our work but in a different task. \citet{Gao2022PALPL} tackles math and symbolic reasoning tasks by decomposing the natural language problem into runnable steps using Code-LLM and delegating solution calculation to a PL interpreter.
We leverage PL features (\eg, inheritance, type annotation) to introduce extra information and constraints for structured prediction, which is largely overlooked by prior work.

\paragraph{Event Extraction}
\citet{li-etal-2013-joint, nguyen-etal-2016-joint-event, yang-mitchell-2016-joint, wadden-etal-2019-entity, lin-etal-2020-joint} use classification models and mitigate error propagation from pipeline models by leveraging global features to predict event triggers and arguments jointly. 
Recent work such as \citet{liu-etal-2020-event} formulates event extraction as a reading comprehension problem and \citet{li-etal-2021-document, huang-etal-2021-document, paolini2021structured, hsu-etal-2022-degree} converts event extraction to a text generation task to better exploit label semantics from pretrained language models. 
The most similar work to ours is Text2Event \cite{lu-etal-2021-text2event}, which uses controlled generation to generate structures in a manually specified linearized format directly, hindering the model in leveraging pre-trained NL knowledge. On the other hand, our approach \CodeForStruct\ directly generates structure in PL instead of using a manually designed format to fully exploit LLM's knowledge of PL.


\section{Conclusions and Future Work}
We propose \CodeForStruct\ for structured prediction tasks in NLP by leveraging LLMs trained on language and code. As a case study, we use \CodeForStruct\ to extract event arguments from natural language sentences through code generation. 
We show that, with sufficient in-context examples, formulating EAE as a code generation problem is advantageous over using text-based prompts.
Our proposed \CodeForStruct\ rivals fully-supervised models trained on 4,202 data instances only using 20-shot. It also outperforms a SOTA model by 29.5\% absolute F1 when both are given the same 20-shot data.
Furthermore, benefitting from hierarchical event definitions, \CodeForStruct~can predict arguments for zero-resource event types only using 10-shot training instances from its sibling event type and outperforms 0-shot baseline by 12\% absolute F1 score.
Going forward, we plan to expand \CodeForStruct~to a broader range of more complex structured prediction tasks (\eg, relation prediction, schema matching). We would further explore the executable nature of PL to improve LLM's ability for structured prediction.

\section*{Limitations}
In this work, our approach assumes event triggers and argument templates (\ie, ontology) are given. 
This limits our approach's applicability, as it requires an event detection system to produce event triggers and event types before LLMs can be prompted to generate event arguments. 

We only explore hierarchical events with only 2 levels from the ACE05-E ontology and data, which has limited coverage of real-world complex event hierarchy.
Similar to prior event argument extraction work, our approach relies on a human-curated hierarchical ontology. We leave automatically discover hierarchical ontology for future work.

Despite LLMs performing well on EAE with few-shot data, compared to existing supervised approaches, their inference is relatively slow and costly\footnote{We perform most of our experiments on \texttt{code-davinci-002} which is in free public beta at the time of the experiment. For \texttt{text-davinci} models, around 700 USD was used to access its API to perform relevant experiments in this paper.} since the LLMs we used are generally more than 100x larger in the number of parameters. 
Prior work \cite{Zhao2021CalibrateBU, lu-etal-2022-fantastically} has demonstrated a strong relationship between performance and in-context demonstrations; however, for ease of comparison to supervised baselines, we use the same set of examples from the training set for in-context learning. We expect better selecting \cite{Liu2021WhatMG} and ordering \cite{lu-etal-2022-fantastically} in-context examples can benefit \CodeForStruct~performance, which we leave for future work.

\section*{Ethical Considerations}

Since event argument extraction only requires predicting arguments from the given text, the risk of generating toxic languages is relatively low as long as the given test is not toxic. This is because the prediction can be grounded in the input sentence, eliminating potential toxic tokens that did not appear in the original sentence.
However, discrimination and bias are possible, as observed in the foundational LLMs we used \cite{gpt3, codex, Ouyang2022TrainingLM}, which we refer to \citet{gpt3} for detailed discussion.

\section*{Acknowledgement}

We thank the anonymous reviewers for their helpful suggestions and comments. This research is based upon work supported by U.S. DARPA KAIROS Program No.
FA8750-19-2-1004, U.S. DARPA AIDA Program No. FA8750-18-2-0014 and U.S. DARPA ITM Program No. FA8650-23-C-7316. 
The views and conclusions contained herein are those of the authors and should not be interpreted as necessarily representing the official policies, either expressed or implied, of DARPA, or the U.S. Government. The U.S. Government is authorized to reproduce and distribute reprints for governmental purposes notwithstanding any copyright annotation therein.

\bibliography{anthology,custom}
\bibliographystyle{acl_natbib}

\newpage
\appendix
\counterwithin{figure}{section}
\counterwithin{table}{section}
\section{Appendix}
\label{sec:appendix}


\subsection{Qualitative Analysis}
\label{sec:qualitative-analysis}

\begin{figure*}[t]
\centering
\includegraphics[width=\textwidth]{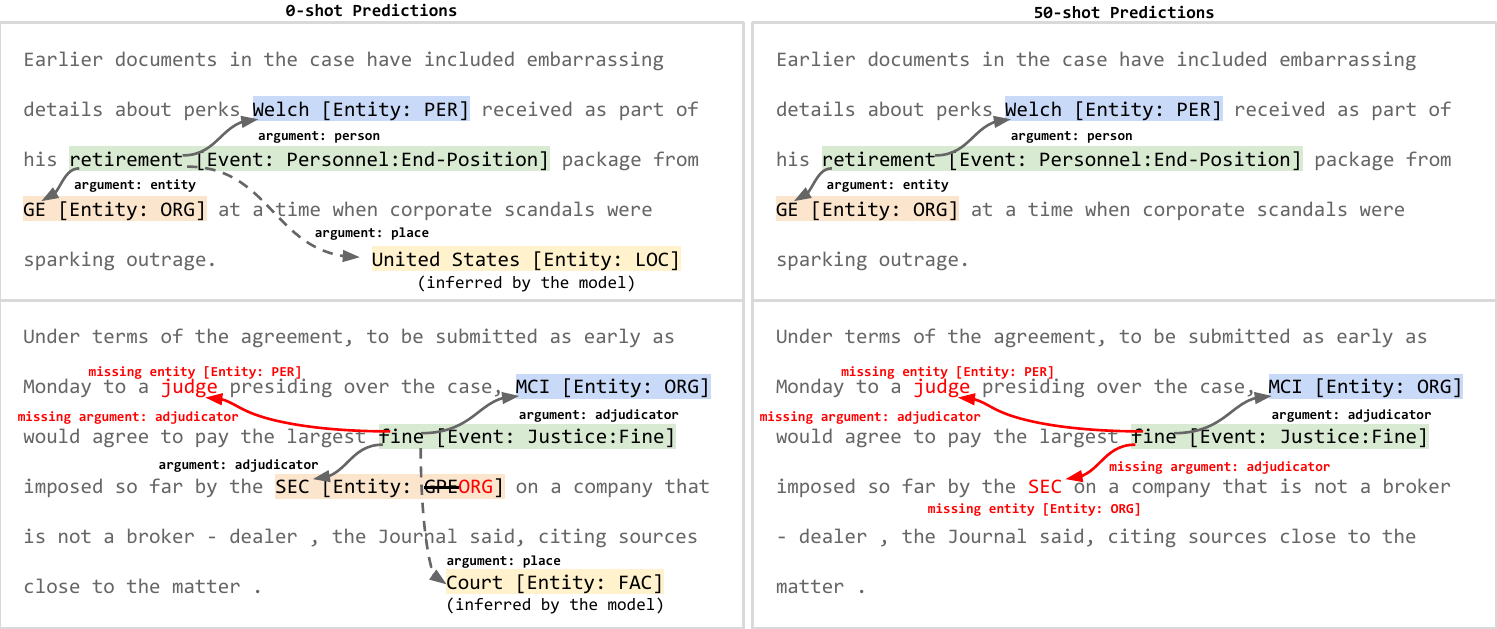}
\caption{Examples of 0-shot and 50-shot \CodeForStruct\ event argument prediction using \texttt{code-davinci-002} on ACE05-E. In both 0-shot examples, LLM can infer an entity that does not present in the text as an argument (marked with a yellow span). \CodeForStruct~predicts fewer arguments when the examples are increased to 50-shot. We mark incorrect predictions with strikethrough text. Entities that LLM failed to predict are marked in red font.
}
\label{fig:qualitative-examples}
\end{figure*}

We show examples of 0-shot and 50-shot \CodeForStruct\ argument extraction result in \fref{fig:qualitative-examples}.
\CodeForStruct\ can leverage implicit commonsense knowledge in LLM to infer arguments not presented in the text. 
In the first 0-shot example, the model inferred the \texttt{place} of Welch's retirement is in the \textit{United States}. This is a reasonable guess since Welch, in this example, is the former CEO of General Electric (GE), whose headquarter is in the United States. In the second 0-shot example, our model inferred that the \texttt{Justice:Fine} event should take \texttt{place} in a \textit{court}, which matches our commonsense knowledge.
Interestingly, we observe that increasing the number of in-context examples from 0-shot to 50-shot inhibits LLM from generating arguments (\ie, making LLMs more conservative), including these inferred arguments and a correctly predicted argument (\ie, \texttt{SEC}) in 0-shot predictions. 
%


\begin{table*}[h!]
\centering
\resizebox{\linewidth}{!}{
\begin{tabular}{l|ccccc|ccccc}
\toprule
{} & \multicolumn{5}{c|}{\textbf{Arg-I F1}} & \multicolumn{5}{c}{\textbf{Arg-C F1}} \\
$k$-shot &    0  &    1  &    10 &    20 &    50 &    0  &    1  &    10 &    20 &    50 \\
\midrule
\CodeForStruct     &  50.6 &  \textbf{57.3} &  57.2 &  \textbf{62.1} &  \textbf{62.3} &  36.0 &  \textbf{47.8} &  52.8 &  \textbf{58.5} &  \textbf{58.1} \\
+ amr              &  51.1 &  54.7 &  55.6 &     - &     - &  \textbf{37.2} &  44.2 &  51.3 &     - &     - \\
+ keywords         &  \textbf{52.3} &  \textbf{57.3} &  \textbf{58.0} &  61.7 &  61.7 &  36.4 &  47.3 &  \textbf{53.5} &  57.7 &  57.9 \\
\bottomrule
\end{tabular}
}
\caption{Prompt components analysis on \texttt{code-davinci-002}. The best scores (in \%) are bolded. - means the result is unavailable due to the input prompt exceeding the corresponding LLM's supported input token length.
}
\label{tab:prompt-component-analysis}
\end{table*}

\subsection{Prompt Component Analysis}
\label{sec:prompt-component-analysis}
In this section, we present an empirical analysis of other prompt component candidates. We compare different prompt components in \tref{tab:prompt-component-analysis} using \texttt{code-davinci-002} and following the same hyper-parameters described in \sref{sec:experiment-setup}.

\begin{figure*}[ht]
\centering
\includegraphics[width=0.9\textwidth]{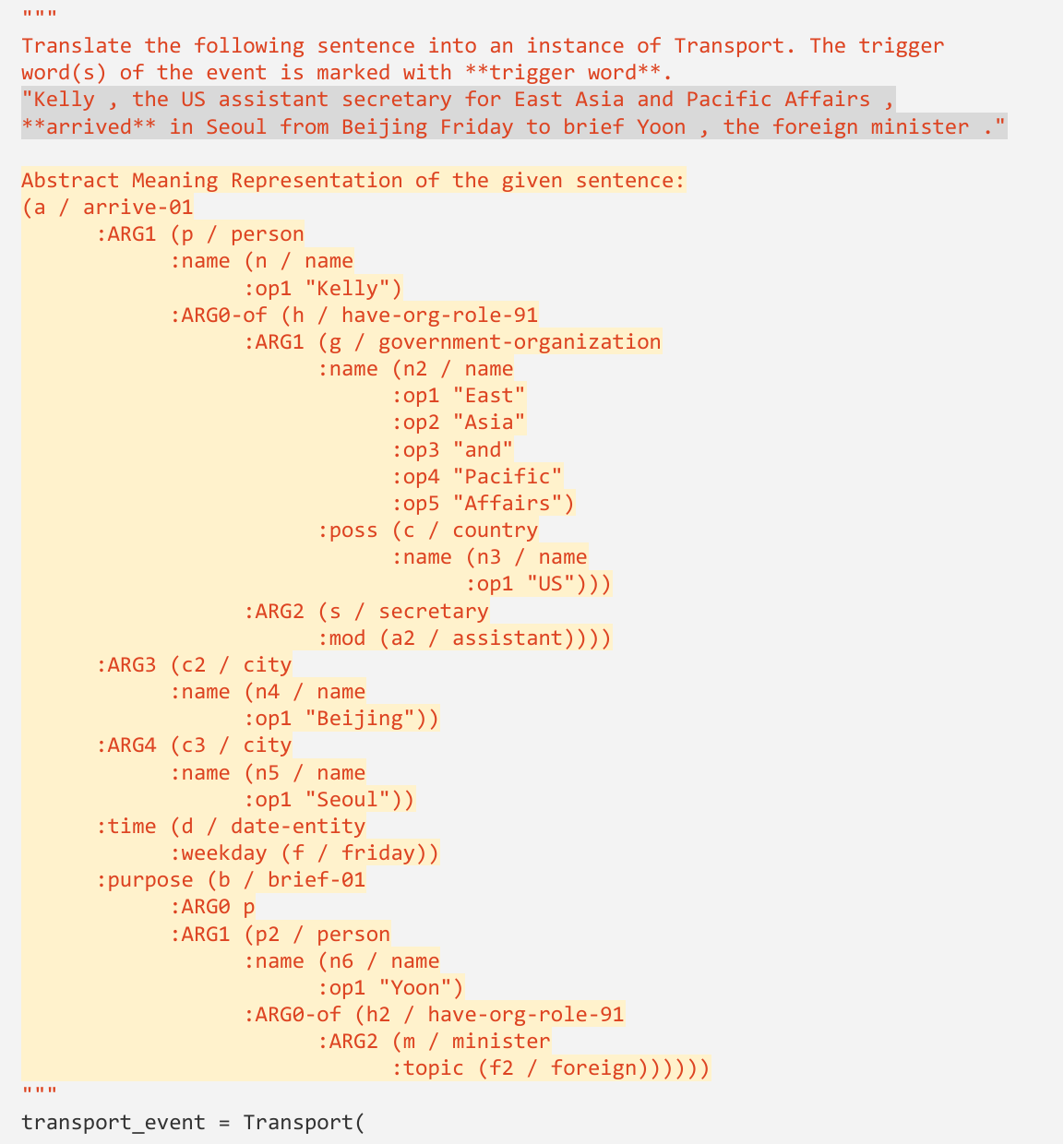}
\caption{Example of an AMR-augmented task prompt. We append the AMR prediction after the input sentence. Different prompt components compared to \CodeForStruct\ are highlighted in yellow. 
}
\label{fig:example-AMR-task-prompt}
\end{figure*}

\begin{figure*}[ht]
\centering
\includegraphics[width=\textwidth]{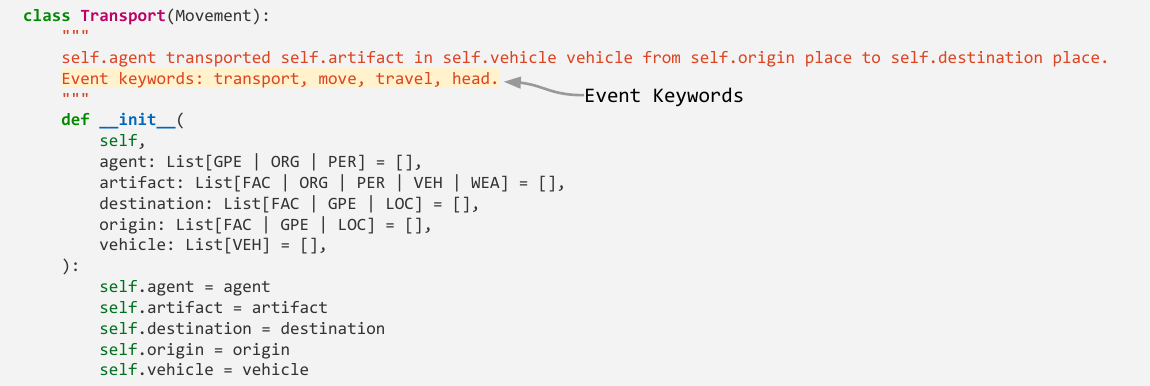}
\caption{Example of an event-keywords-augmented event definition. Different prompt components compared to \CodeForStruct\ are highlighted in yellow. We use event keywords from \citet{li-etal-2021-document}.
}
\label{fig:example-keywords-event-definition}
\end{figure*}

\begin{itemize}
    \item \textbf{Event Keywords} We augment event-related keywords into the docstring of event definition for \CodeForStruct\ (illustrated in \fref{fig:example-keywords-event-definition}). We follow the same keywords used by \citet{li-etal-2021-document}. 
    
    \item \textbf{AMR} \citet{zhang-ji-2021-abstract} have demonstrated the effectiveness of utilizing Abstract Meaning Representation (AMR) \cite{banarescu-etal-2013-abstract} for information extraction. We experiment with AMR-augmented prompts. We use \texttt{armlib} \footnote{\href{https://github.com/bjascob/amrlib}{https://github.com/bjascob/amrlib}, parse\_xfm\_bart\_large v0.1.0} to predict AMR, and append the AMR structure after the NL sentence in the task prompt \sref{sec:task-prompt} (see \fref{fig:example-AMR-task-prompt} for an example). 
\end{itemize}

Prompts that include event keywords and AMR all perform slightly better than \CodeForStruct\ under the zero-shot setting on all metrics (\tref{tab:prompt-component-analysis}).

\begin{table*}[h!]
\centering
\resizebox{\linewidth}{!}{
\begin{tabular}{l|ccccc|ccccc}
\toprule
{} & \multicolumn{5}{c|}{\textbf{Arg-I F1}} & \multicolumn{5}{c}{\textbf{Arg-C F1}} \\
$k$-shot &    0  &    1  &    10 &    20 &    50 &    0  &    1  &    10 &    20 &    50 \\
\midrule
\CodeForStruct     &  50.6 &  \textbf{57.3} &  57.2 &  \textbf{62.1} & \textbf{62.3} &  36.0 &  47.8 &  52.8 &  \textbf{58.5} & \textbf{58.1} \\
- trigger         &  48.8 &  54.4 &  53.0 &  57.6 &  56.6 &  33.8 &  44.1 &  48.9 &  53.8 &  51.5 \\
- description     &  \textbf{51.4} &  56.7 &  56.2 &  61.1 &  61.6 &  \textbf{36.1} &  47.2 &  51.6 &  57.1 &  57.8 \\
- type annotation &  49.4 &  57.2 &  \textbf{58.0} &  61.5 &  61.4 &  35.7 &  \textbf{48.0} &  \textbf{54.5} &  57.6 &  57.5 \\
- hierarchy       &  49.4 &  56.6 &  55.5 &  59.9 &  60.4 &  34.3 &  46.8 &  50.0 &  55.4 &  55.9 \\
\bottomrule
\end{tabular}
}
\caption{Ablation study on \texttt{code-davinci-002}. The best scores (in \%) are bolded. - means the result is unavailable due to the input prompt exceeding the corresponding LLM's supported input token length. 
}
\label{tab:ablation-study}
\end{table*}

\subsection{Ablation Study}
\label{sec:ablation-study}

In \tref{tab:ablation-study}, we ablate different prompt components described in \sref{sec:prompt-construction}, including event trigger marking, event description in natural language, type annotation, and hierarchical ontology. We perform this ablation study using \texttt{code-davinci-002}.

\paragraph{Event Trigger Marking} We find that removing event trigger marking consistently degrades performance on all metrics over varying numbers of in-context examples.

\paragraph{Event Description} Event descriptions generally provide a small F1 gain under the few-shot setting.
However, removing event descriptions improves \CodeForStruct's zero-shot performance on argument identification. 0-shot Arg-I precision is relatively unchanged after removing event descriptions (37.4 vs. 37.2). We argue that removing event descriptions loosens entity-related constraints and allows LLM to identify more relevant entities. This is supported by the improvement of 0-shot Arg-I recall (78.7 to 81.8) after description removal, which mainly accounts for the increase in 0-shot Arg-I F1. Despite being helpful in argument identification by boosting 0-shot Arg-I recall, we do not see the benefit of removing descriptions in few-shot Arg-C, where it performs consistently worse compared to \CodeForStruct.

\paragraph{Type Annotation} Type annotation is more helpful when more in-context examples are provided ($k \geq 20$). Under a low-shot setting, the F1 difference resulting from type annotation removal is small and inconsistent across different shots. Prompts with type annotation consistently outperforms prompts without it when sufficient in-context examples are provided ($k \geq 20$). We hypothesize that type annotations help disambiguate entity types accepted for each argument, and such disambiguation ability is only needed when the number of entity instances that appeared in in-context examples passes a certain threshold (\eg, $k \geq 20$).

\paragraph{Hierarchical Event Definition} Providing hierarchical event definition (i.e., the parent class definition of a given child event class) benefits \CodeForStruct\ performance in high-shot setting ($k \geq 20$). Prompts without parent class definition perform on par with \CodeForStruct\ under $k < 20$.

\subsection{In-context Example Variability Analysis}
\label{sec:variability-analysis}

\begin{figure}[h]
\centering
\resizebox{\columnwidth}{!}{
\includegraphics[]{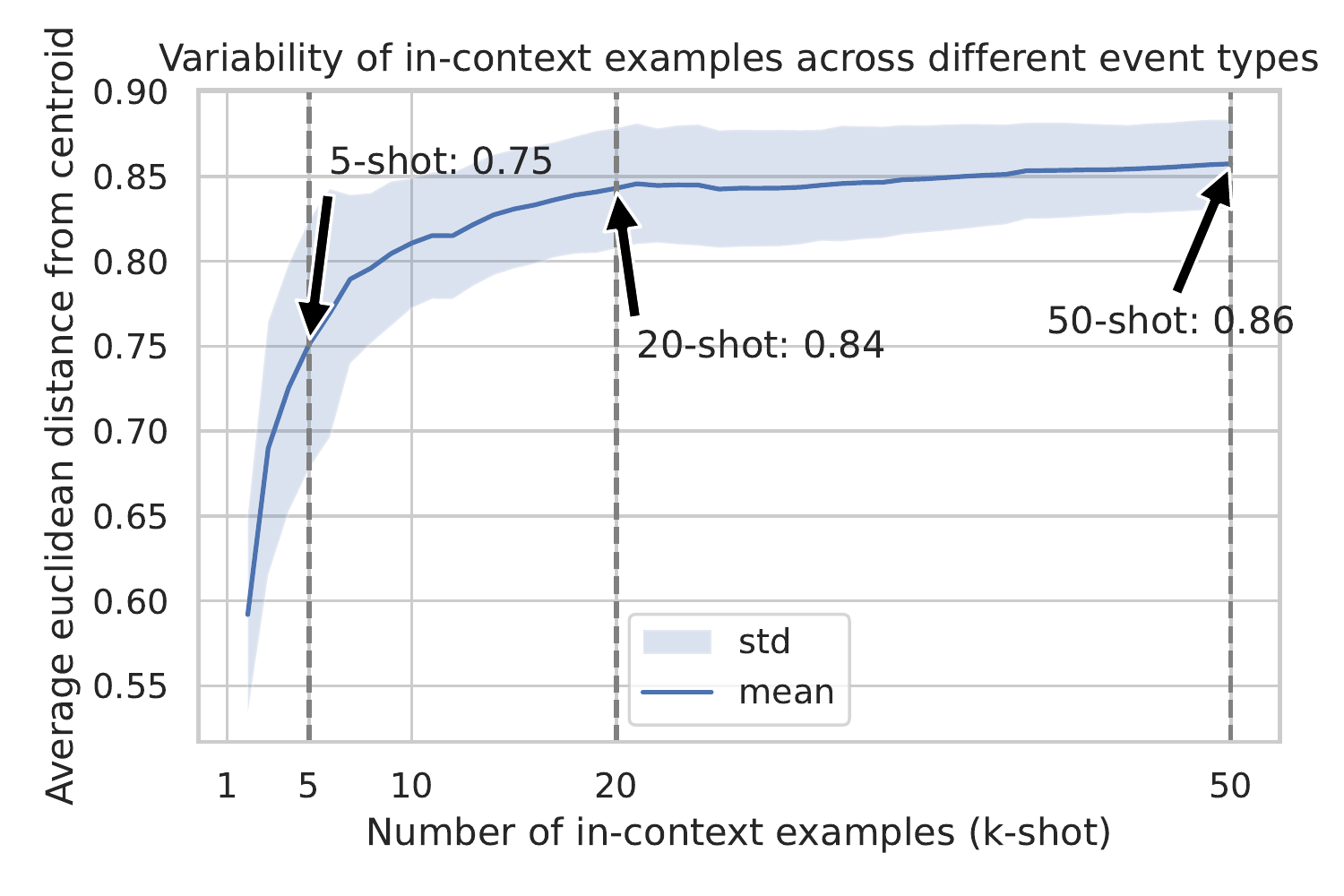}
}
\caption{
The variability of $k$-shot in-context examples (\ie, average euclidean distance from centroid example) across different event types increases with diminishing returns when $k$ increases.
}
\label{fig:variability-vs-k}
\end{figure}

\begin{figure}[h]
\centering
\resizebox{\columnwidth}{!}{
\includegraphics[]{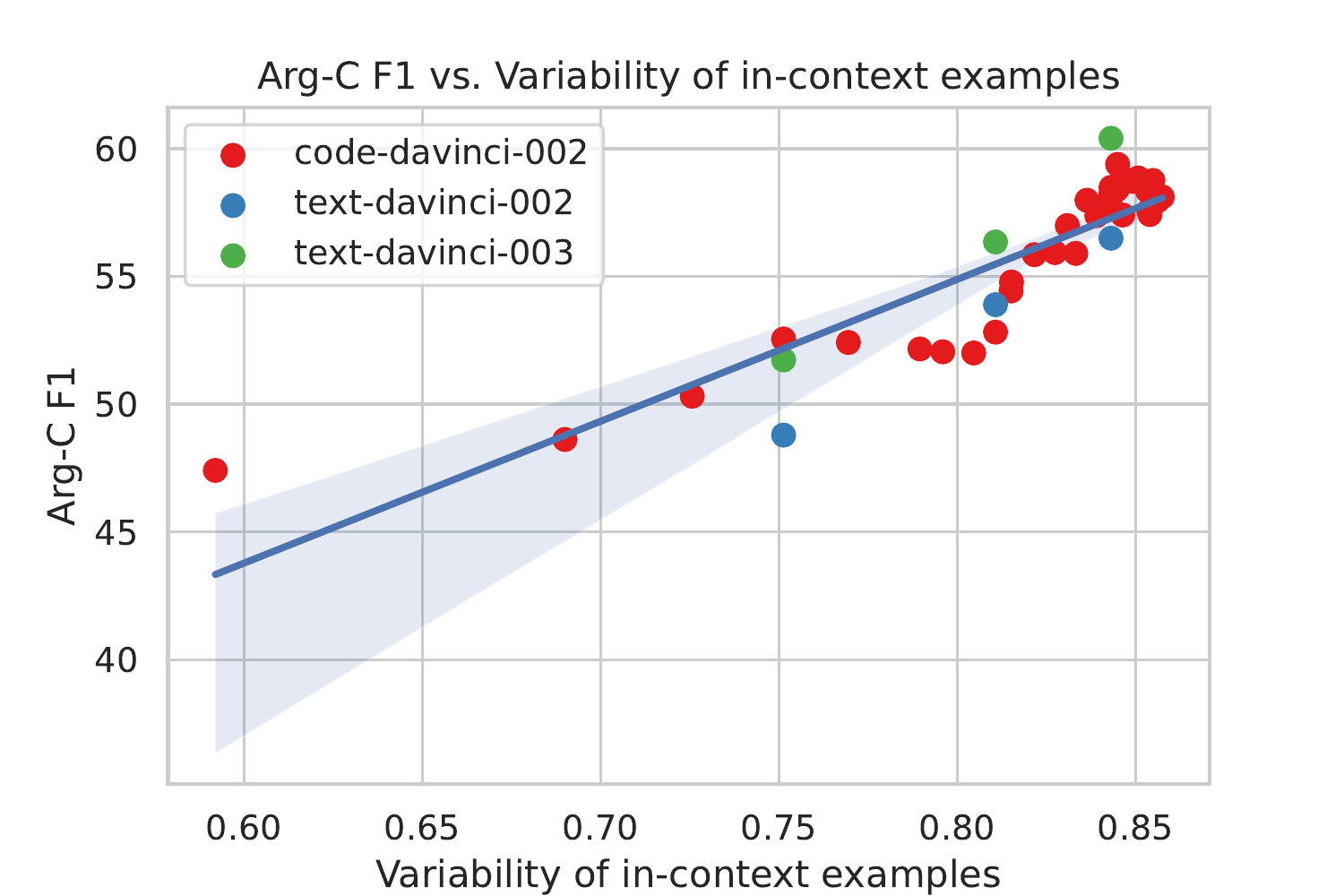}
}
\caption{
The variability of in-context examples is positively correlated with code prompt Arg-C performance.
}
\label{fig:arg-c-vs-variability}
\end{figure}

To investigate why the Arg-C performance plateaus with an increasing number of in-context examples $k$ as shown in \fref{fig:model-performance}, we analyze the sentence variability of a fixed set of in-context examples (\sref{sec:in-context-learning}).
We consider the set of $k$-shot in-context examples for each event type $e$ as a cluster $D_{e}$ where $|D_{e}| \leq k$ and use \texttt{sentence-transformer}\footnote{\texttt{all-mpnet-base-v2} model} to embed all the input sentences from $D_{e}$ into a cluster of vectors $V_{e}$.

We use the average euclidean distance from the centroid example similar to \cite{Halkidi2001OnCV} to measure the variability of in-context examples for each event type $e$: 
$$
\text{Variability}(e) = \frac{1}{|V_{e}|}\sum_{v \in V_{e}} d(v, \bar{v})
$$
where $d(\cdot, \cdot)$ is the euclidean distance between two vectors and $\bar{v} = \frac{1}{|V_{e}|}\sum_{v \in V_{e}} v $ is the centroid example of the cluster $V_{e}$.

We calculate the mean $\text{Variability}(e)$ across all $e$ for $k \in \{1, \cdots, 50\}$.
In \fref{fig:variability-vs-k}, similar to Arg-C performance in \fref{fig:model-performance}, we find the mean $\text{Variability}(e)$ across all $e$ increases with diminishing returns with increasing $k$. Furthermore, we find that, in \fref{fig:arg-c-vs-variability}, Arg-C F1 performance is positively correlated with the mean $\text{Variability}(e)$ across all $e$. This suggests the lack of in-context example variability improvement could be one of the reasons Arg-C F1 plateaus, even when more examples are given.

\begin{figure*}[bh]
\centering
\includegraphics[width=\textwidth]{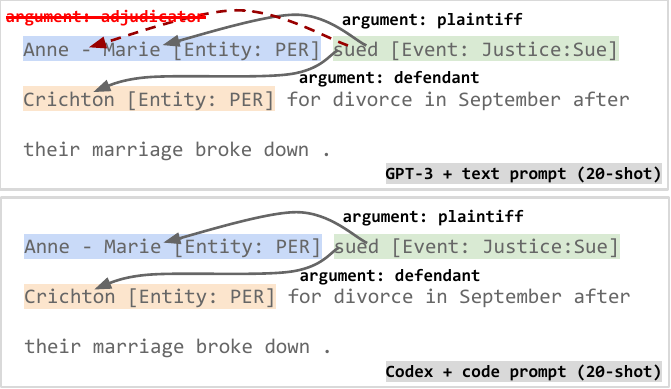}
\caption{Example prediction of 20-shot text prompt $T^{(1)}$ using \texttt{text-davinci-002} and code prompt using \texttt{code-davinci-002}. In this example, 20-shot text prompt using \texttt{text-davinci-002} incorrectly predicts the same entity \textit{Anne-Marie} as both \textit{adjudicator} and \textit{plaintiff} of the \texttt{Justice:Sue} event.
}
\label{fig:code-vs-text-prompt-qualitative-example}
\end{figure*}

\begin{table*}[htbp]
\centering
\resizebox{\textwidth}{!}{%

\begin{tabular}{lr|cc|cc|cc}
\toprule
     & Model & \multicolumn{2}{c}{code-davinci-002} & \multicolumn{2}{c}{text-davinci-002} & \multicolumn{2}{c}{text-davinci-003} \\
     & Metric &            Arg-I & Arg-C &            Arg-I & Arg-C &            Arg-I & Arg-C \\
Prompt & $k$-shot &                  &       &                  &       &                  &       \\
\midrule
\multirow{7}{*}{code} & 0  &             50.6 &  36.0 &             48.9 &  35.0 &             49.9 &  37.8 \\
     & 1  &             57.3 &  47.8 &             55.8 &  45.2 &             56.0 &  44.7 \\
     & 5  &             58.0 &  52.5 &             56.0 &  48.8 &             59.2 &  51.7 \\
     & 10 &             57.2 &  52.8 &             60.6 &  53.9 &             62.8 &  56.3 \\
     & 20 &             62.1 &  58.5 &             59.9 &  56.5 &             65.0 &  60.4 \\
     & 30 &             62.2 &  58.4 &                - &     - &                - &     - \\
     & 50 &             62.3 &  58.1 &                - &     - &                - &     - \\
\hline

\multirow{5}{*}{\shortstack{text (code imitation)}} & 0  &             49.9 &  38.2 &             51.5 &  37.4 &             52.0 &  39.2 \\
     & 1  &             57.2 &  48.8 &             54.0 &  42.2 &             57.5 &  47.9 \\
     & 5  &             56.9 &  49.6 &             58.0 &  45.8 &             60.1 &  50.3 \\
     & 10 &             58.6 &  52.0 &             57.9 &  47.5 &             59.7 &  51.3 \\
     & 20 &             60.4 &  54.9 &             59.0 &  48.5 &             61.5 &  52.6 \\
\hline

\multirow{5}{*}{\shortstack{text (BART-Gen style \citet{li-etal-2021-document})}} & 0  &              0.0 &   0.0 &             28.7 &  21.9 &             34.6 &  25.2 \\
     & 1  &             52.6 &  43.1 &             50.5 &  40.3 &             54.9 &  43.6 \\
     & 5  &             56.1 &  51.4 &             55.0 &  47.4 &             59.9 &  53.8 \\
     & 10 &             57.4 &  52.7 &             57.7 &  48.9 &             62.2 &  57.5 \\
     & 20 &             61.9 &  56.1 &             56.2 &  50.7 &             64.3 &  60.8 \\
\bottomrule
\end{tabular}
}
\caption{Performance of the code and two variants of the text prompts on the Arg-I and Arg-C metrics. 50-shot results for \texttt{text-davinci} and text prompts are unavailable since the 50-shot prompt length exceeds such LLM's input token limitation.
Examples of text prompt variants can be found in \fref{fig:IE-puretext-gen} (code imitation) and \fref{fig:IE-naturaltext-gen} (BART-Gen style).
}
\label{tab:text-and-code-prompt-perf-numbers}
\end{table*}

\begin{figure*}[h!]
\centering
\includegraphics[width=\textwidth]{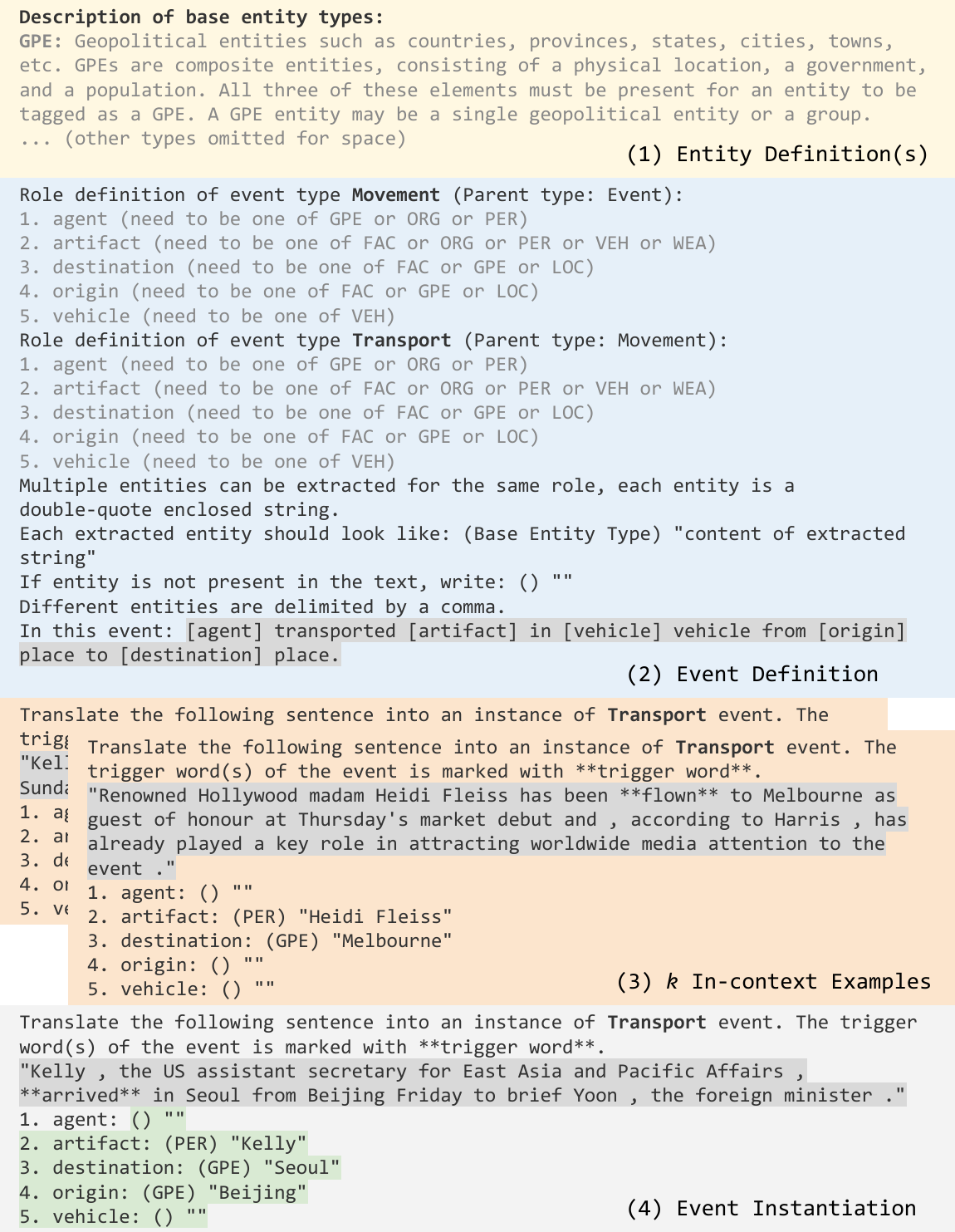}
\caption{Natural language prompt for EAE task following our code prompt design described in section \ref{sec:prompt-construction}.
We ask a LLM to generate event instantiation marked in green.
}
\label{fig:IE-puretext-gen}
\end{figure*}

\begin{figure*}[h!]
\centering
\includegraphics[width=\textwidth]{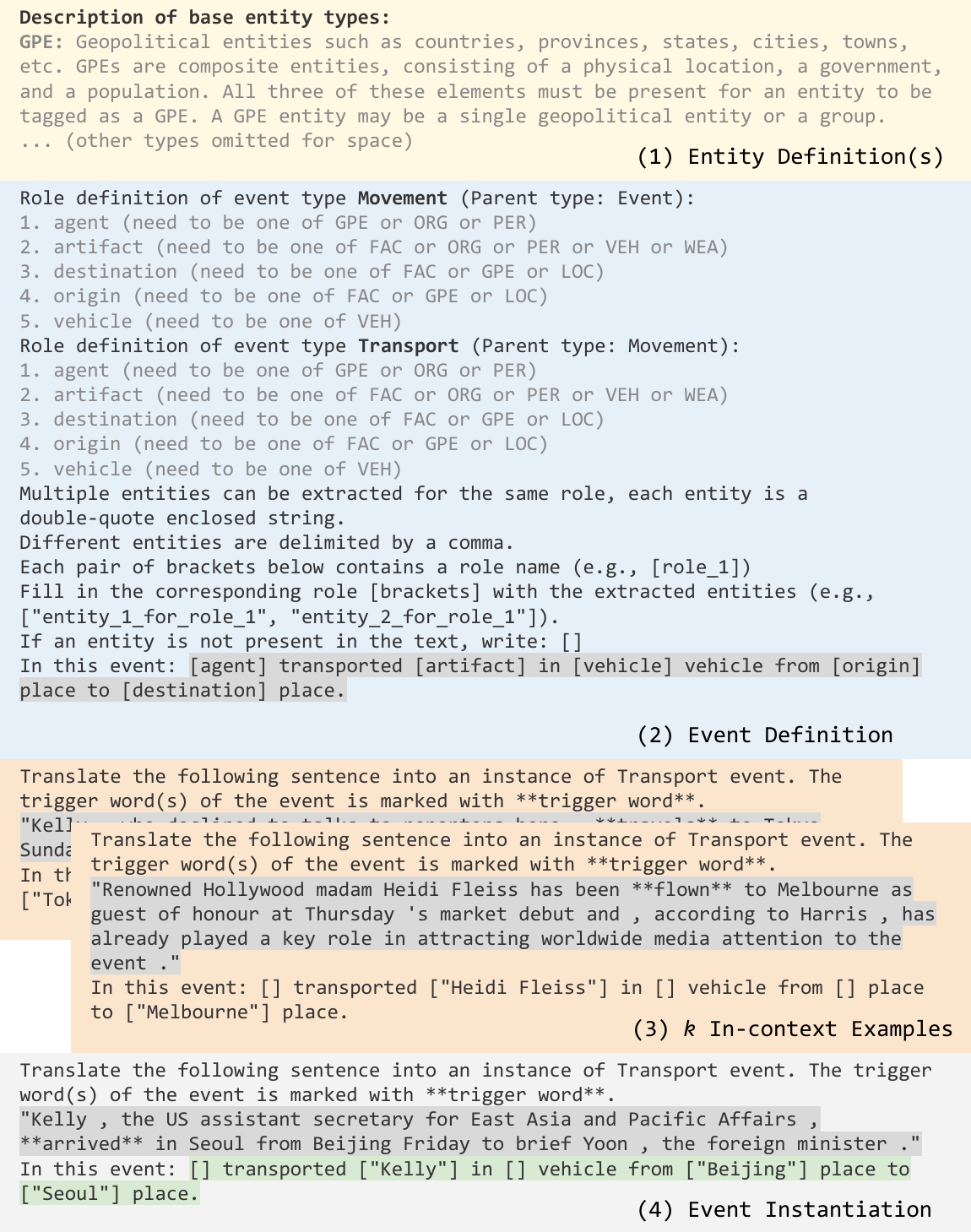}
\caption{BART-Gen style \cite{li-etal-2021-document} natural language prompt for EAE task.
We ask a LLM to generate event instantiation marked in green. 
Brackets and double-enclosed strings are designed for ease of parsing free form natural language.
}
\label{fig:IE-naturaltext-gen}
\end{figure*}

\begin{figure*}[ht]
\centering
\includegraphics[width=\textwidth]{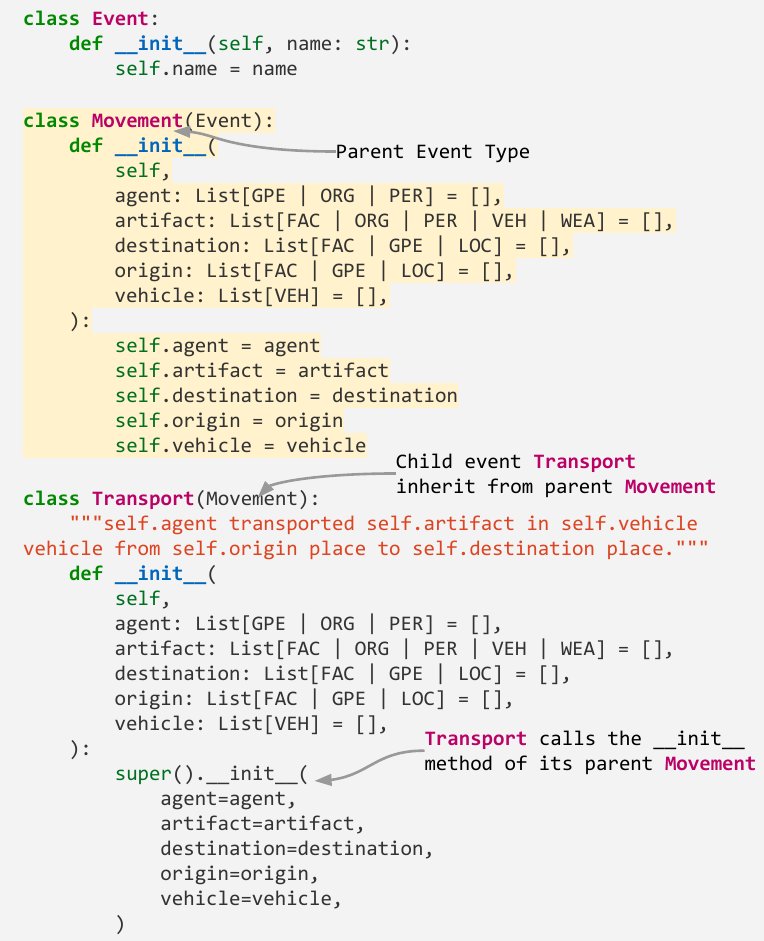}
\caption{Example of a hierarchical event definition. Different prompt components compared to \CodeForStruct\ are highlighted in yellow.}
\label{fig:example-hierarchical-event-definition}
\end{figure*}

\begin{figure*}[ht]
\centering
\includegraphics[width=0.85\textwidth]{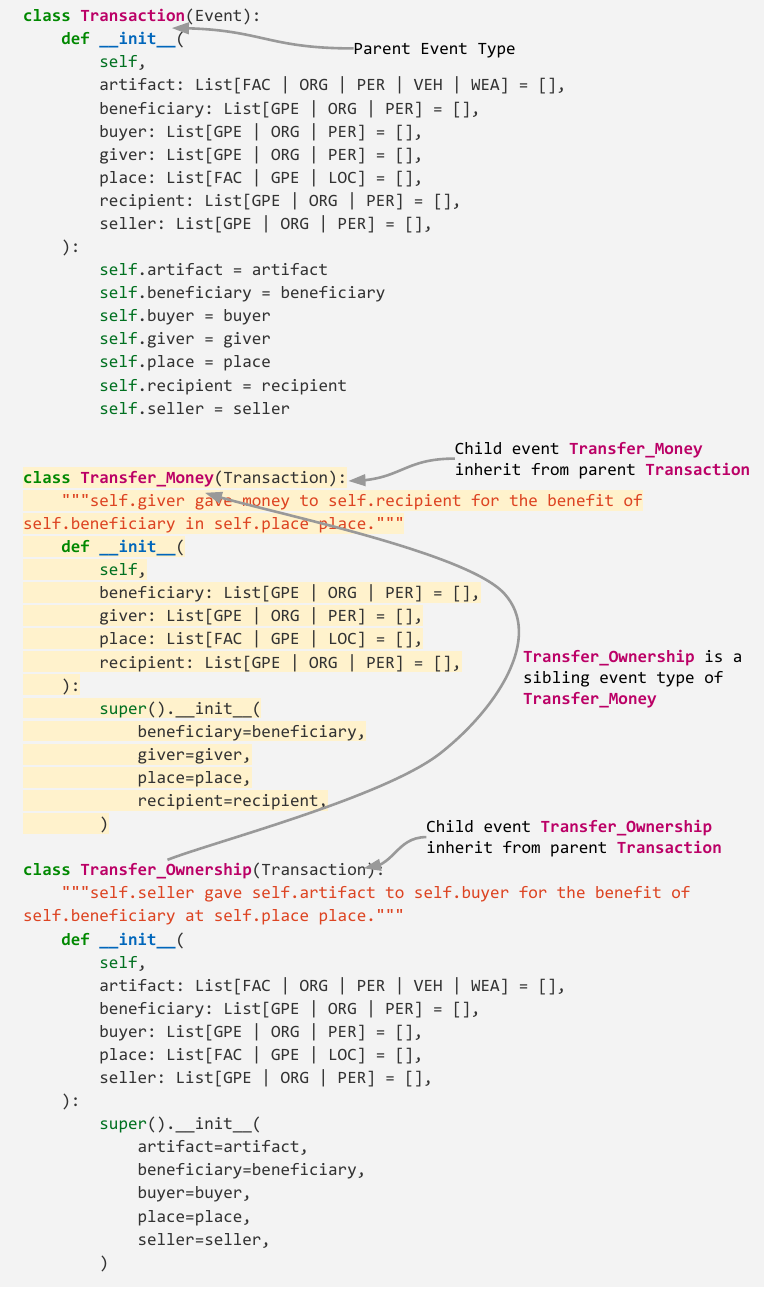}
\caption{Example of a hierarchical event definition with a sibling event type. Different prompt components compared to \fref{fig:example-hierarchical-event-definition} are highlighted in yellow.}
\label{fig:example-hierarchical-sibling-event-definition}
\end{figure*}

\begin{figure*}[ht]
\centering
\includegraphics[width=0.9\textwidth]{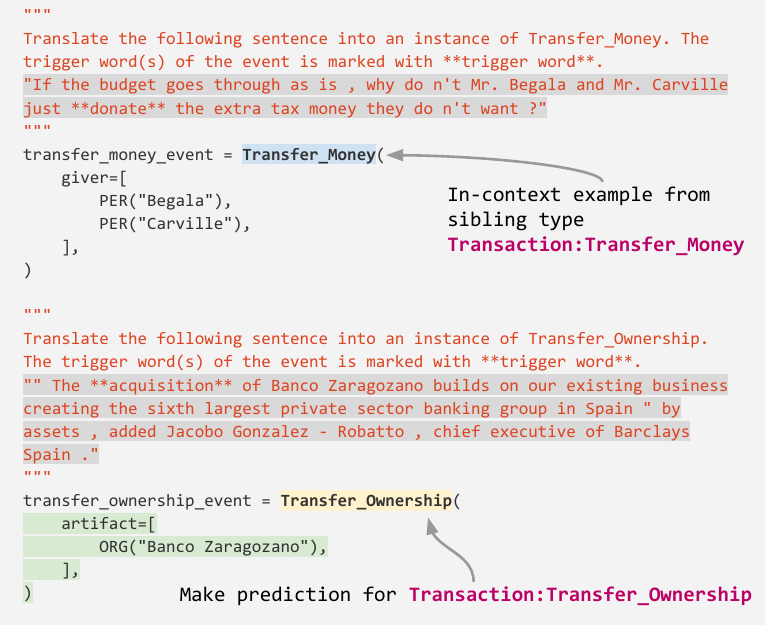}
\caption{Example of a task prompt with a 1-shot example from sibling event type. Event definitions for the task prompt is shown in \fref{fig:example-hierarchical-sibling-event-definition}. Groundtruth prediction is colored green.
}
\label{fig:example-hierarchical-sibling-event-example}
\end{figure*}

\begin{table*}[th]
\centering
\begin{tabular}{llrr}
\toprule
&& \# of Test Instances  &  \# of Train Example \\
Parent Event Type & Child Event Type &             &              \\
\midrule
\multirow{4}{*}{Business} & Declare-Bankruptcy &           2 &           39 \\
            & End-Org &           5 &           24 \\
            & Merge-Org &           0 &           13 \\
            & Start-Org &          17 &           21 \\
\hline
\multirow{2}{*}{Conflict} & Attack &          90 &         1211 \\
            & Demonstrate &           7 &           62 \\
\hline
\multirow{2}{*}{Contact} & Meet &          49 &          194 \\
            & Phone-Write &           8 &          104 \\
\hline
\multirow{13}{*}{Justice} & Acquit &           1 &            4 \\
            & Appeal &           6 &           30 \\
            & Arrest-Jail &           6 &           72 \\
            & Charge-Indict &           8 &           95 \\
            & Convict &           6 &           61 \\
            & Execute &           2 &           12 \\
            & Extradite &           1 &            6 \\
            & Fine &           6 &           22 \\
            & Pardon &           0 &            2 \\
            & Release-Parole &           1 &           44 \\
            & Sentence &          11 &           83 \\
            & Sue &           4 &           60 \\
            & Trial-Hearing &           5 &          103 \\
\hline
\multirow{5}{*}{Life} & Be-Born &           3 &           44 \\
            & Die &          17 &          516 \\
            & Divorce &           9 &           20 \\
            & Injure &           1 &          125 \\
            & Marry &          10 &           71 \\
\hline
Movement & Transport &          47 &          561 \\
\hline
\multirow{4}{*}{Personnel} & Elect &          13 &          156 \\
            & End-Position &          17 &          143 \\
            & Nominate &           1 &           11 \\
            & Start-Position &          11 &           87 \\
\hline
\multirow{2}{*}{Transaction} & Transfer-Money &          12 &          121 \\
            & Transfer-Ownership &          27 &           85 \\
\bottomrule
\end{tabular}

\caption{The number of Train/Test event instances for 33 event types in ACE05-E.}
\label{tab:event-instance-stats}
\end{table*}

\begin{table*}[th]
\centering
\begin{tabular}{lll}
\toprule
            &                &                    \\
Parent Event Type & Child Event Type (Train) & Child Event Type (Test) \\
\midrule
\multirow{3}{*}{Business} & \multirow{3}{*}{Declare-Bankruptcy} & End-Org \\
            &                & Merge-Org* \\
            &                & Start-Org \\ \hline
Conflict & Attack & Demonstrate \\ \hline
Contact & Meet & Phone-Write \\ \hline
\multirow{12}{*}{Justice} & \multirow{12}{*}{Trial-Hearing} & Acquit \\
            &                & Appeal \\
            &                & Arrest-Jail \\
            &                & Charge-Indict \\
            &                & Convict \\
            &                & Execute \\
            &                & Extradite \\
            &                & Fine \\
            &                & Pardon* \\
            &                & Release-Parole \\
            &                & Sentence \\
            &                & Sue \\
\hline
\multirow{4}{*}{Life} & \multirow{4}{*}{Die} & Be-Born \\
            &                & Divorce \\
            &                & Injure \\
            &                & Marry \\
\hline
\multirow{3}{*}{Personnel} & \multirow{3}{*}{Elect} & End-Position \\
            &                & Nominate \\
            &                & Start-Position \\
\hline
Transaction & Transfer-Money & Transfer-Ownership \\
\bottomrule
\end{tabular}

\caption{Train/Test split for each parent event type. * denotes child event types that do not have examples in the ACE05-E test set.
}
\label{tab:event-type-hierarchy-train-test-split}
\end{table*}

\end{document}